\pdfoutput=1

\documentclass[11pt]{article}

\usepackage{authblk}
\usepackage[]{acl2023}

\usepackage{times}
\usepackage{latexsym}
\usepackage{graphicx}
\usepackage{float}
\usepackage{stfloats}
\usepackage{multirow}
\usepackage{amsfonts} 
\usepackage{amsmath} 
\usepackage{booktabs}
\usepackage{cleveref}
\crefformat{section}{\S#2#1#3}
\crefformat{subsection}{\S#2#1#3}

\usepackage[T1]{fontenc}

\usepackage[utf8]{inputenc}

\usepackage{microtype}

\usepackage{inconsolata}

\title{Learning to Generalize for Cross-domain QA}

\author{
    \textbf{Yingjie Niu$^*$ \textsuperscript{\rm1,2}},
    \textbf{Linyi Yang$^*$ \textsuperscript{\rm3,4}},
    \textbf{Ruihai Dong\textsuperscript{\rm1,2}}, 
    \textbf{Yue Zhang \textsuperscript{\rm3,4} }
    \\

    \textsuperscript{1} School of Computer Science, University College Dublin\\
    \textsuperscript{2} SFI Centre for Research Training in Machine Learning \\
    \textsuperscript{3} Institute of Advanced Technology, Westlake Institute for Advanced Study \\
    \textsuperscript{4} School of Engineering, Westlake University \\
    
    \texttt{\{yingjie.niu\}@ucdconnect.ie}, \texttt{\{ruihai.dong\}@ucd.ie}\\
    \texttt{\{yanglinyi, zhangyue\}@westlake.edu.cn}\\
    
}

\begin{document}
\maketitle
\def\thefootnote{*}\footnotetext{These authors contributed equally to this work.}
\begin{abstract}
There have been growing concerns regarding the out-of-domain generalization ability of natural language processing (NLP) models, particularly in question-answering (QA) tasks. Current synthesized data augmentation methods for QA are hampered by increased training costs. To address this issue, we propose a novel approach that combines prompting methods and linear probing then fine-tuning strategy, which does not entail additional cost. Our method has been theoretically and empirically shown to be effective in enhancing the generalization ability of both generative and discriminative models. Our approach outperforms state-of-the-art baselines, with an average increase in F1 score of 4.5\%-7.9\%. Furthermore, our method can be easily integrated into any pre-trained models and offers a promising solution to the under-explored cross-domain QA task. We release our source code at Github\footnote{\url{https://github.com/FreddieNIU/Prompt-QA}}.
\end{abstract}

\section{Introduction}

Question answering (QA) models \cite{why2016question,trischler-etal-2017-newsqa,lewis2021question,gu2021beyond} aim to answer passage-based questions automatically with the help of facts in a given context (sometimes referred to as machine reading comprehension \cite{dua2019drop,sen-saffari-2020-models}). Over the last few years, pre-trained models have achieved great progress on a variety of large-scale datasets, e.g., SQuAD \cite{rajpurkar2016squad}, NewsQA \cite{trischler-etal-2017-newsqa}, DROP \cite{dua2019drop}, CoRA \cite{asai2021question}, and NarrativeQA \cite{Kocisky2017narrativeqa}. However, existing methods can suffer significant performance degradation when the tuned system is directly applied to out-of-domain examples \cite{gururangan2018annotation,wu2019understanding,tripuraneni2020theory,ICLR20Counterfact,malinin2021shifts,varshney2022investigating}. 

This paper focuses on a novel cross-domain QA task where we assume models trained on the source domain can be generalized to the target domain, where no labeled or unlabeled data is available. As shown in Figure 1, QA pairs from different domains have intrinsically different feature distributions. For example, in the technology field, the context can frequently contain ``e-commerce'' and ``network''. While in the pharmaceutical sector, the context can consist of ``COVID-19'', ``vaccine'', and ``diagnostic'' more frequently. Cross-domain QA poses significant challenges to real-world scenarios, and it is proved that even large-scale pre-trained models \cite{gu2021beyond} can still encounter performance degradation under domain generalization.


\begin{figure}[t]
    \centering
    \includegraphics[width=.95\linewidth]{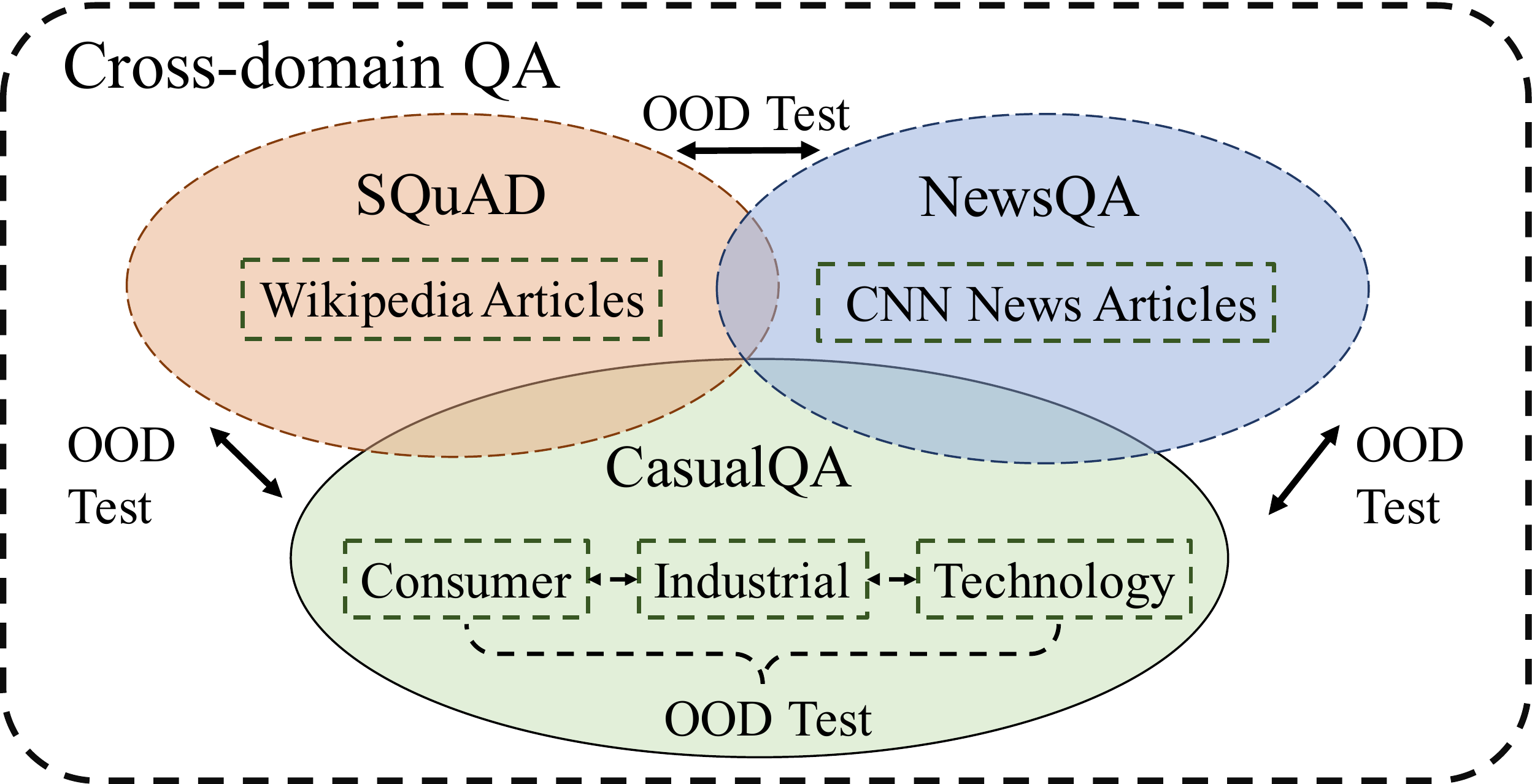}
    \caption{Cross-domain QA task consists of three datasets from different domains.}
    \label{fig:1}
\end{figure}

To address these drawbacks, we introduce a novel cross-domain QA setting, focusing on the methods that consistently improve the domain generalization performance without additional computational costs. Intuitively, cross-domain QA can benefit from prompting in which instances from different domains can share a unified set of label words. Thus, no additional parameters can carry domain-specific information to hinder the OOD generalization for an unseen domain. However, using the prompt method solely could increase the risk of overfitting and bring limited benefits, as prompt templates are fixed, which may be learned as spurious features by models. Thus, we consider using the linear-probing and then fine-tuning (LP-FT) strategy to reduce the reliance between prompt patterns with labels by freezing pre-trained parameters. In this way, LP-FT can benefit cross-domain QA by preventing pre-trained features from being distorted when tuning on a specific domain \cite{kumar2022finetuning}. Prompting-based LP-FT method does not introduce new parameters, so the performance decay when training on a source domain and testing on a new target domain can be reduced without entailing additional cost. 

Under the LP-FT framework, we introduce four prompt types to extract invariant features in different domains: question type, sentiment, named entity, and key phrase. These prompts aim to increase the question similarity and benefit the model in generalizing to out-of-domain questions. Existing prompting methods have not been applied to natural language processing tasks beyond simple fine-tuning settings. To enable promoting methods to adapt LP-FT, we theoretically prove that LP-FT still holds consistently better robustness for prompting methods (Section 3.3).

We experiment on three different domain datasets (Figure 1). Results show that our prompt-based LP-FT method significantly improves the performance of cross-domain QA models on either the standard hold-out or OOD tests, with an average increase in F1 of 4.5\%-7.9\% compared to baselines. Also, our method consistently outperforms the standard fine-tuning strategy on both discriminative and generative models. Besides, we provide an in-depth analysis of the ablation study towards the OOD robustness that details the efficacy of LP-FT and prompting methods, respectively. To our knowledge, we are the first to present a new zero-shot cross-domain QA task and propose a novel Prompt-based LP-FT method. All resources are available at \url{https://github.com/FreddieNIU/Prompt-QA}.

\section{Related Work}
\textbf{Out-of-domain} performance degradation has attracted considerable research interest recently. A line of work \cite{morgan2015counterfactuals,wang2021robustness,ICLR21Counterfact,yang2021exploring,malkiel2021maximal,lu2022rationale} aims to improve models' generalization ability on text classification. Differently, we investigate the OOD generalization problem on the QA task. 

\citet{lewis2021question} and \citet{wang2021can} find that 60-70\% of test-time answers of popular open-domain QA benchmark datasets exist in the training set, and it is proved that training set memory plays a vital role in testing. \citet{liu2021challenges} empirically prove that language models suffer performance degradation when there is no train-test set overlapping. To test the actual generalization ability of QA models, several novel QA datasets have been constructed and released, focusing on evaluating QA models on out-of-domain generalization ability \cite{gu2021beyond}. \citet{yang2022towards} present the first cross-domain QA dataset and observe a performance decay problem regarding the OOD test. Many existing methods intend to improve the OOD performance of QA models through data augmentation. \citet{yue2022synthetic} introduce a synthesizing question-answer pairs method to improve target-domain QA performance. In contrast, we propose a prompt-based method combined with linear probing and then fine-tuning, which is more computationally efficient and does not require target domain annotations.

\begin{figure*}[t]
    \centering
    \includegraphics[width=.95\linewidth]{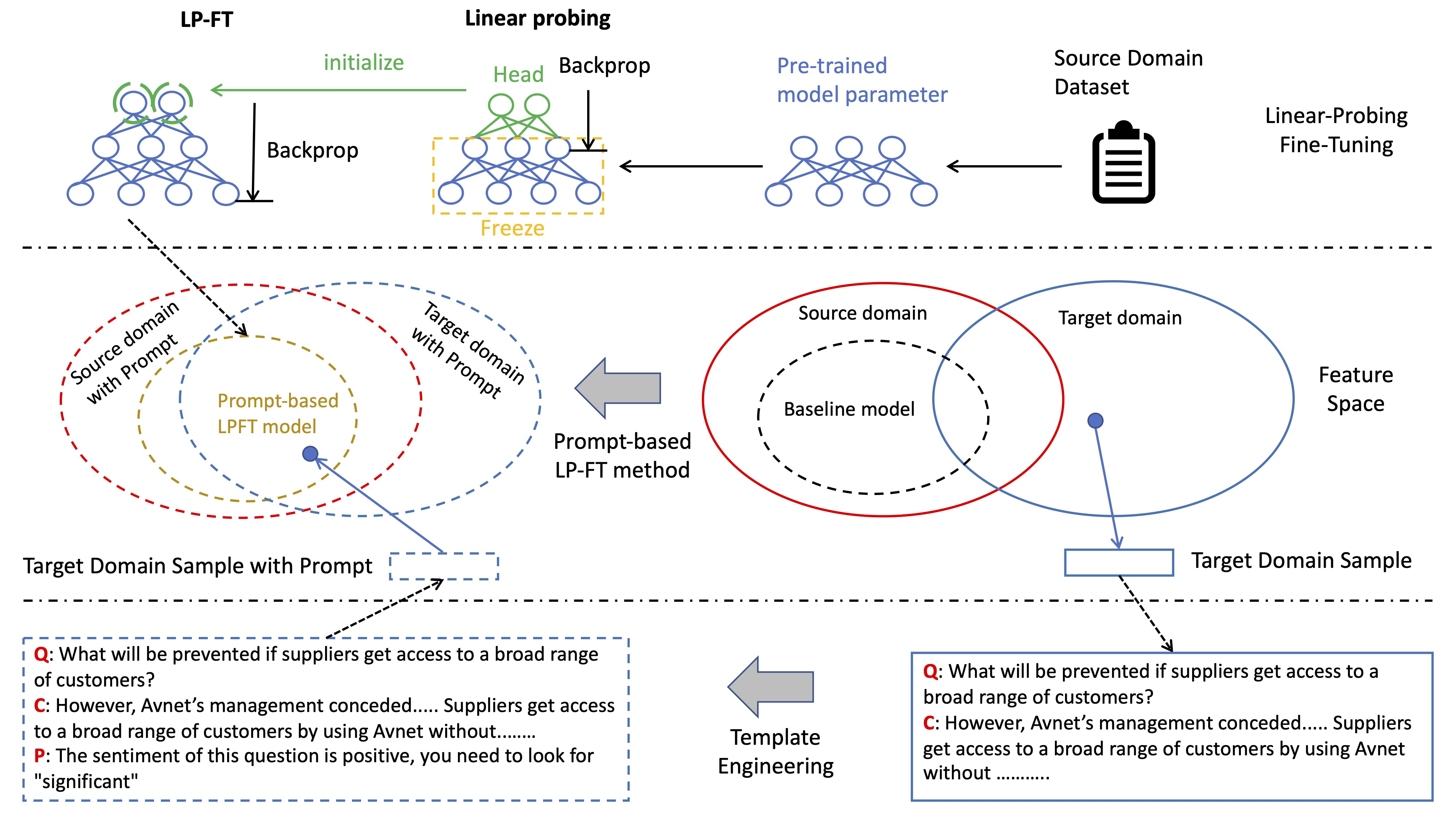}
    \caption{The workflow of the prompt-based linear-probing then fine-tuning strategy. The \textit{bottom} part shows the template engineering process where we add a prompt for each sample. The \textit{middle} part represents the feature space of the Prompt LP-FT model and baseline models. Compared with the baseline model, in the feature space, the prompt-based LPFT model is superior in two respects: the distance between the feature distributions of the two domains is reduced, and the features learned by the model are closer to the intersection of the two domains. The \textit{top} part demonstrates the linear probing and then fine-tuning process. The ``Q'', ``C'' and ``P'' represent ``Question'', ``Context'' and ``Prompt'' in a sample, respectively (\textit{in bottom})}. 
    \label{fig:summary}
\end{figure*}

\textbf{Prompt-based methods} on pre-trained language models have received considerable research interest. The paradigm "\textit{pre-train, prompt, and predict}" replaces the "\textit{pre-train, fine-tune}" procedure for improved few-shot learning \cite{liu2021promptsurvey}. Prompt-based methods have been applied not only in sentence-level few-shot learning tasks, such as named entity recognition \cite{ma2021templateFree} but also in sophisticated learning tasks like natural language understanding \cite{wang2022ACL_promDA}. However, little work applies prompts on the cross-domain QA tasks \cite{jacovi2021contrastive}. We leverage the fixed-format characteristic of the prompt to extract the invariant features in the changing dataset to enhance the OOD generalization of the model. 

Instead of fine-tuning, \emph{linear probing} is an alternative to tuning a pre-trained model on a downstream task. \citet{liu2019linearprobing,tripuraneni2020theory} examine the representations produced by several recent pre-trained language models and find that linear models trained on top of frozen contextual representations are competitive with state-of-the-art task-specific fine-tuned models in many cases but fail in tasks where fine-grained language knowledge is required. \citet{kumar2022finetuning} theoretically prove that the linear-probing then fine-tuning (LP-FT) approach can enhance the OOD generalization ability of pre-trained models. In our work, we are the first to provide theoretical evidence that LP-FT still holds consistently better robustness for prompting methods in NLP.

\section{Method}

Figure \ref{fig:summary} illustrates the workflow of the Prompt-based LP-FT method. We first generate a prompt for each input sample through template engineering and prompt designing (\cref{sec:3-1}). Then, the source domain dataset with prompts is used for linear probing and then fine-tuning (\cref{sec:LPFT}) a pre-trained model. (\textit{top}).
Compared with the baseline model, in the feature space, the prompt-based LPFT model is superior in two respects: the distance between the feature distributions of the two domains is reduced, and the features learned by the model are closer to the intersection of the two domains (\textit{middle}). The feature space demonstrates how the prompt-based method and the LP-FT strategy benefit the cross-domain QA, respectively, and also shows the motivation for using prompt-based LP-FT to benefit the cross-domain QA task.

\subsection{Template Designing}\label{sec:3-1}
\begin{figure}[t]
    \centering
    \includegraphics[width=\linewidth]{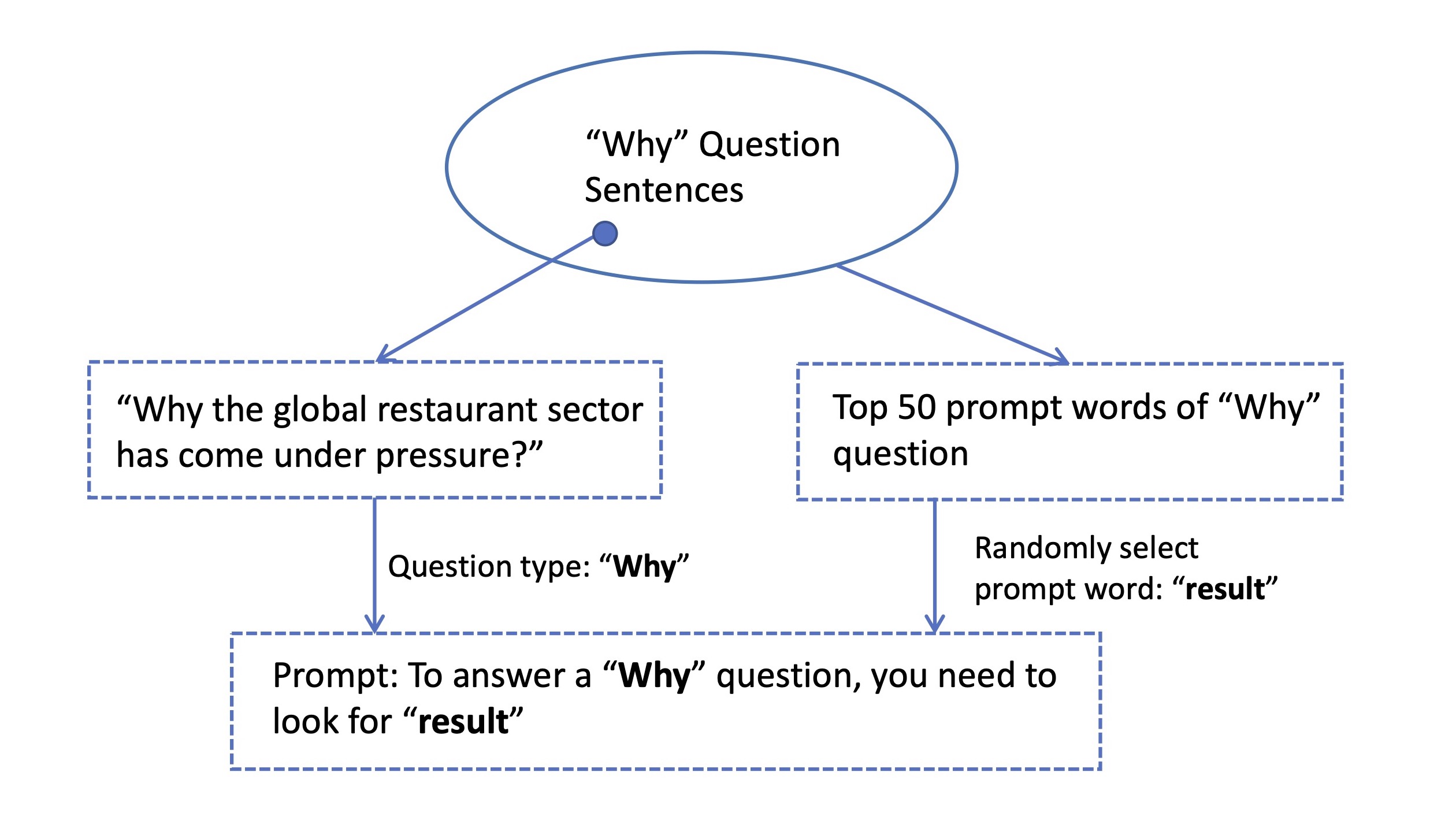}
    \caption{Generating question-type prompts}
    \label{fig:gen_qtype}
\end{figure}

We take a \textit{Template Engineering} process to look for a prompt template that results in the most effective performance on a given task. The template designing rules can be found in \emph{\cref{sec:template-engineering}}. The prompt design is inspired by the process of a non-native speaker (or a non-professional reader) reading articles (or professional documents) and answering questions. They may lack some depth of knowledge, such as the meanings of less commonly used words (or domain-specific knowledge). Language models may encounter similar situations in the cross-domain QA task. We design four types of templates. Figure \ref{fig:gen_qtype} gives an example of a question-type template. Other template designs can be found in \emph{\cref{sec:template-designing}}. Below, we take the question-type template as an example to illustrate the template designing process: 

\textbf{Question-type Templates.} 
Suppose that for a given question, \emph{``Why have we increased our projections for cancer drug Loxo305 and diabetes drug tripeptide is useful?''}, a human tester tries to find the answer from the article. In the question, users might not understand tokens such as \emph{Loxo305, diabetes, tripeptide}, etc. However, if the user is aware that the question might be about ``Why something is useful?", then she/he can search some keywords such as \emph{because}, \emph{as}, and \emph{since} from the article and the context following these words, which might help her/him to find the correct answer. For each type of question, some specific words might help to locate their answers.


We consider four typical types of questions. For each question type, we first find out the most related words with it, such as \emph{because}, \emph{since} with the question type \emph{why}, by measuring Pointwise Mutual Information (PMI) scores \cite{bouma2009normalized} between candidate words and the question type. Afterward, we select the 50 most related words to generate a prompt for each question. 




\textbf{Loss Functions.} For a prompt-based QA task, given a question-context-prompt tuple $(Q, C, P)$, we calculate the probability of each word $ c_n $ being the start position or end position of the correct answer for discriminative models as follows:
\begin{equation}
\small
p(c_n |Q,C,P) = \emph{Softmax}(W_{head}h_{c_n} + b_{head})
\end{equation} where $ Q \in \mathbb{R}^{s_{q} \times d_{word} } $, $ C \in \mathbb{R}^{s_{c} \times d_{word} } $, and $ P \in \mathbb{R}^{s_{p} \times d_{word} } $ denote the question, context and prompt, respectively. $ s_{q}/s_{c}/s_{p}$ and $ d_{word} $ denote the \emph{question/context/prompt sentence length} and the \emph{word embedding dimension}, respectively. 
$\ h_{c_n} $ denotes the feature representation of $(Q, C, P)$ concatenated on the first dimension produced by a pre-trained model, $ W_{head} \in \mathbb{R}^{\nu \times d_h } $ and $ \ b_{head} \in \mathbb{R}^{ \nu } $. $d_h$ denotes the dimension of $ h_{c_n} $ and $ \nu $ denotes the length of answer sentence. The loss function is the sum of the cross entropy for start and end positions. 

\begin{equation}
    \mathcal{L}_{dis} = - \sum \limits_{n=1}^m \log p(c_n|Q,C,P)
\end{equation}
where $m$ is the number of words in $C$.

We regard the QA task as a Seq2Seq generation task for generative models and use the LM loss, 
\begin{equation}
    \mathcal{L}_{gen} = - \sum \limits_{n} \log p(c_n|c_{<n},Q,C,P)
\end{equation}
where $c_{<n}$ denotes the generated words.

\subsection{Linear Probing then Fine-tuning}\label{sec:LPFT}
 
The OOD generalization problem is defined as follows \cite{kumar2022finetuning}: given a predictor $f$: to map inputs $X $ to outputs $Y$. For some loss function $\mathcal{L}$, the predictor in-domain performance $L_{id}$ and out-of-domain performance $L_{ood}$ are:
 
\begin{equation}
\begin{aligned}
    L_{id}(f) = \mathop{\mathbb{E}}\limits_{(X,\textbf{y})\sim p_{id}}[\mathcal{L}(f(X)), \textbf{y}] \\
    L_{ood}(f) = \mathop{\mathbb{E}}\limits_{(X,\textbf{y})\sim p_{ood}}[\mathcal{L}(f(X)), \textbf{y}] 
\end{aligned}
\end{equation}
where the predictor is evaluated on test samples $(X,\textbf{y})$ drawn from in-domain distribution $P_{id}$, and also evaluated on test samples $(X,\textbf{y})$ drawn from out-of-domain distribution $P_{ood}$. To simplify the formula representation, in this paper, $X$ represents the question, and context ($Q$ and $C$); $\textbf{y}$ indicates the answer sentence.
 
The final predictor $f$ is parameterized as a feature extractor and a linear "head". Hence, the training loss is:
\begin{equation}
    \hat{\mathcal{L}}(\textbf{v},B) = ||XB^\top \textbf{v}-\textbf{y}||^2_2
    \label{equ:5}
\end{equation}
where $\textbf{v}$ denotes the linear head and $B$ denotes the feature extractor. We assume that the initial feature extractor $B_0$ is obtained from the pre-trained model, considering two methods to learn a predictor $f_{\textbf{v}, B}$: 1) linear probing where $B = B_0$ and the linear head is obtained by minimizing some loss on the training data \cite{liu2019linearprobing}, and 2) fine-tuning where both $\textbf{v}$ and $B$ are updated on the training data with $B$ initialized as $B_0$ \cite{kumar2022finetuning}. 

\subsection{Theoretical Proof}
We prove that linear probing and then fine-tuning improves the results for prompt tuning by extending the proof for standard fine-tuning \cite{kumar2022finetuning}. In particular, the derivative of Eq \ref{equ:5} with respect to the feature extractor $B$ is:
\begin{equation}
    \nabla_B\hat{\mathcal{L}}(\textbf{v},B) = 2\textbf{v}(\textbf{y}-XB^\top \textbf{v})^\top X
    \label{equ:6}
\end{equation}

For Eq \ref{equ:6}, if $U$ is a sample extracted from a direction orthogonal subspace to the training subspace, $\nabla_B\hat{\mathcal{L}}(\textbf{v},B)U = 0$,  the training process on $X$ will not decrease the loss on the orthogonal subspace. However, the gradient is not zero for directions in the ID subspace. This explains why fine-tuning can achieve a higher ID performance but a lower OOD performance.

In our proposed prompt-based method, the prompt $P$ is concatenated to the original $X$ (along the sentence length dimension), and the equation can be expressed below:
\begin{equation}
    \nabla_B\hat{\mathcal{L}}_p (\textbf{v},B) = 2\textbf{v}(Y-(X+P)B^\top \textbf{v})^\top (X+P)
\end{equation} where assume that $P$ is not orthogonal to the $X$ or its orthogonal subspace. Consequently, we have $\nabla_B\hat{\mathcal{L}_p}(\textbf{v},B)(U+P) \neq 0$. In this way, the training process on $X$ with prompt $P$ would modify the loss on the OOD samples with the prompt.

In the linear probing and then fine-tuning method, the OOD error of fine-tuning is 
\begin{equation}
    \sqrt{L_{ood}(\textbf{v}_{ft},B_{ft}(t))} \geq \sigma \frac{min(\varphi, \varphi^2/||w_*||_2)}{(1+||w_*||_2)^2}
\end{equation}
where $\textbf{v}_{ft}$ and $B_{ft}$ are the linear head and feature extractor after fine-tuning. $\sigma$ is a fixed parameter \cite{kumar2022finetuning} related to $B_0$. $w_* = \textbf{v}_*B_*$, $\textbf{v}_*$ and $B_*$ are the optimal parameters. $\varphi$ is the initial head alignment error $\varphi = |(\textbf{v}_0^\top \textbf{v}_*)^2-(\textbf{v}_*^\top \textbf{v}_*)^2|$. In order to decrease the OOD error, the head $\textbf{v}_{0}$ has to be as close to the $\textbf{v}_*$ as possible. It is proved that initializing the head with $\textbf{v}_{lp}$ (LP-FT) can decrease the OOD error \cite{kumar2022finetuning} more than random initializing head with $\textbf{v}_0$ (FT) since $\textbf{v}_0$ is far away from $\textbf{v}_*$. Converting input $X$ to $X+P$ does not affect $\frac{min(\varphi, \varphi^2/||w_*||_2)}{(1+||w_*||_2)^2}$, implying the LP-FT strategy can be applied after we introduce $P$. As a result, the Prompt-based LP-FT strategy is used to avoid distorting pre-trained features.

\section{Experimental Setup}
We introduce experiments' datasets, baseline methods, and evaluation metrics in this section. \subsection{Datasets}

We evaluate the proposed method on three datasets: \textbf{Causal QA} \cite{yang2022towards}, \textbf{SQuAD 1.1} \cite{weissenborn-etal-2017-making} and \textbf{NewsQA} \cite{trischler-etal-2017-newsqa}. All datasets are in English. Domain-related information is provided in the CausalQA dataset, which is valuable for cross-domain question-answering tasks. For the in-domain test, we experiment on the whole CausalQA dataset before splitting into domains(domain-independent QA) and on each particular domain after splitting into domains. The distribution change and word overlap between datasets can be found in \cref{sec:word-overlap}.

For the OOD test, we have two experiment setups: \textbf{Setup 1)} we split the CausalQA dataset, based on the domain information, into mutually exclusive training/validation/testing sets in the same ratio of 8:1:1. \textbf{Setup 2)} we conduct OOD tests across different datasets from different domains. In cross-domain QA, both the training and validation sets of the source domain are used in the training process for hyperparameter searching. The testing sets of source and target domains are used for in-domain evaluation and OOD tests, respectively.  



\subsection{Baseline Models}


Based on the novel cross-domain QA setting, we establish baselines using generative models -- BART \cite{Lewis2019bart}, T5 \cite{raffel2019exploring} -- and discriminative models -- BERT \cite{devlin2018bert}, RoBERTa \cite{liu2019roberta}, and SpanBERT \cite{joshi2020spanbert} with the help of Huggingface framework \cite{wolf2020transformers}. We also implement the commonly used domain adaptation method in previous works \cite{yue2022synthetic, cao2020unsupervised} to compare with our method. The AdamW optimizer has a default learning rate of $10^{-5}$. Other hyper-parameters are tuned by optimizing the performance on the validation set. The standard fine-tuning strategy is considered a baseline when compared to our methods by using four strategies:


\begin{enumerate}
    \item Baseline: we select the RoBERTa-base model as the baseline of discriminative methods.
    \item Baseline + P: we adopt the same baseline models and fine-tuning strategy, only replacing the original dataset with the prompted dataset.
    \item Baseline + LP-FT: we first tune the last linear layer (the ``Head'' for question answering) parameters and replace the head parameters initialized by Huggingface framework models with the head parameters after linear probing. The original dataset is used in this section. 
    \item Baseline + P + LP-FT: the LP-FT strategy is adopted on the dataset with the prompt.
\end{enumerate}

\begin{table*}[t]
\resizebox{\linewidth}{!}{
\begin{tabular}{llcccccc|llllllll}
\hline
\multicolumn{8}{l|}{\textbf{BART}}                                                                                                                                                                                                                                                 & \multicolumn{8}{l}{\textbf{RoBERTa}}                                                                                                                                                                        \\ \hline
\multicolumn{2}{l|}{\multirow{2}{*}{\textbf{Train/Test}}}                       & \multicolumn{2}{c}{\textbf{Consumer}}                     & \multicolumn{2}{c}{\textbf{Industrial}}                          & \multicolumn{2}{c|}{\textbf{Technology}}                          & \multicolumn{2}{c|}{\multirow{2}{*}{\textbf{Train/Test}}}                       & \multicolumn{2}{l}{\textbf{Consumer}} & \multicolumn{2}{l}{\textbf{Industrial}} & \multicolumn{2}{l}{\textbf{Technology}} \\
\multicolumn{2}{l|}{}                                                           & F1                          & EM                          & F1                                 & EM                          & F1                                 & EM                           & \multicolumn{2}{c|}{}                                                           & F1                    & EM            & F1                     & EM             & F1                     & EM             \\ \hline
\multicolumn{1}{l|}{\multirow{2}{*}{\textbf{Con}}}  & \multicolumn{1}{l|}{Ori}  & \textbf{70.29}              & 24.53                       & 68.44                              & 24.07                       & 68.61                              & 23.61                        & \multicolumn{1}{c|}{\multirow{2}{*}{\textbf{Con}}}  & \multicolumn{1}{c|}{Ori}  & \textbf{78.20}        & 51.38         & 72.49                  & 47.68          & 74.63                  & 49.07          \\ \cline{2-2} \cline{10-10}
\multicolumn{1}{l|}{}                               & \multicolumn{1}{l|}{Ours} & (+0.36)                     & (+0.47)                     & (+1.7)                             & (+2.78)                     & (+3.39)                            & (+4.63)                      & \multicolumn{1}{c|}{}                               & \multicolumn{1}{c|}{Ours} & (+0.19)               & (+1.86)       & (+2.58)                & (+0.93)        & (+2.83)                & (+0.47)        \\ \cline{1-2} \cline{9-10}
\multicolumn{1}{l|}{\multirow{2}{*}{\textbf{Ind}}}  & \multicolumn{1}{l|}{Ori}  & \multicolumn{1}{l}{70.11}   & \multicolumn{1}{l}{31.31}   & \multicolumn{1}{l}{\textbf{72.53}} & \multicolumn{1}{l}{32.41}   & \multicolumn{1}{l}{69.63}          & \multicolumn{1}{l|}{27.27}   & \multicolumn{1}{l|}{\multirow{2}{*}{\textbf{Ind}}}  & \multicolumn{1}{l|}{Ori}  & 77.81                 & 49.45         & \textbf{80.05}         & 58.46          & 77.91                  & 48.35          \\ \cline{2-2} \cline{10-10}
\multicolumn{1}{l|}{}                               & \multicolumn{1}{l|}{Ours} & \multicolumn{1}{l}{(+3.95)} & \multicolumn{1}{l}{(+4.75)} & \multicolumn{1}{l}{(+4.05)}        & \multicolumn{1}{l}{(+7.48)} & \multicolumn{1}{l}{(+3.84)}        & \multicolumn{1}{l|}{(+9.93)} & \multicolumn{1}{l|}{}                               & \multicolumn{1}{l|}{Ours} & (+2.74)               & (+9.57)       & (+0.65)                & (+2.20)        & (+1.09)                & (+9.02)        \\ \cline{1-2} \cline{9-10}
\multicolumn{1}{l|}{\multirow{2}{*}{\textbf{Tech}}} & \multicolumn{1}{l|}{Ori}  & \multicolumn{1}{l}{69.89}   & \multicolumn{1}{l}{30.30}   & \multicolumn{1}{l}{69.53}          & \multicolumn{1}{l}{27.77}   & \multicolumn{1}{l}{\textbf{71.79}} & \multicolumn{1}{l|}{33.83}   & \multicolumn{1}{l|}{\multirow{2}{*}{\textbf{Tech}}} & \multicolumn{1}{l|}{Ori}  & 75.54                 & 55.05         & 73.99                  & 48.98          & \textbf{76.49}         & 54.04          \\ \cline{2-2} \cline{10-10}
\multicolumn{1}{l|}{}                               & \multicolumn{1}{l|}{Ours} & (+2.78)                     & (+3.03)                     & (+2.98)                            & (+3.45)                     & (+2.23)                            & (+0.51)                      & \multicolumn{1}{l|}{}                               & \multicolumn{1}{c|}{Ours} & (+2.39)               & (+0.00)       & (+2.71)                & (+7.08)        & (+0.27)                & (-0.50)        \\ \hline
\end{tabular}}
\centering
\caption{Out-of-domain test results of the BART-base model (Left) and RoBERTa-base (Right) on CausalQA (Setup 1). The numbers in brackets represent the performance improved by our method. "Ori" denotes the original fine-tuning method, and ``Ours'' denotes the Prompt-based LP-FT method.}
\label{tab:2}
\end{table*}

\begin{table}[ht]
\centering
\small
\begin{tabular}{lllll}
\hline
\multicolumn{1}{c}{\textbf{}}  & \multicolumn{2}{c}{\textbf{Dev}}         & \multicolumn{2}{c}{\textbf{Test}}        \\ 
\textbf{Methods}               &  F1             & EM             & F1             & EM             \\ \hline
\multicolumn{1}{l|}{BART}                      & 74.16          & 36.50           & 73.26          & 34.49          \\ 
\multicolumn{1}{l|}{BART + LP-FT}                & 74.06          & 35.03          & 73.83          & 34/00             \\ 
\multicolumn{1}{l|}{BART + P}             & 75.60           & 37.47          & 75.33        & 37.66 \\ 
\multicolumn{1}{l|}{BART + P + LP-FT}     &             \textbf{77.60}  & \textbf{41.22} & \textbf{76.90} & \textbf{39.44}\\ \hline
\multicolumn{1}{l|}{RoBERTa}                   & 83.97          & 61.82 & 83.45          & 61.28          \\ 
\multicolumn{1}{l|}{RoBERTa + LP-FT}            & 84.80           & 62.15          & 83.49          & 61.18          \\ 
\multicolumn{1}{l|}{RoBERTa + P}          & 84.55          & \textbf{62.20}           & 83.61          & 61.34          \\ 
\multicolumn{1}{l|}{RoBERTa + P + LP-FT}   & \textbf{84.56} & 62.15          & \textbf{83.87} & \textbf{61.42} \\ \hline
\end{tabular}
\caption{Results of Domain-Independent QA on CausalQA dataset. 'F1' refers to Macro F1, EM refers to exact match. The model name refers to base models, "+P" denotes the base model+prompt method, "+LP-FT" denotes the base model+LP-FT method. }
\label{tab:1}
\end{table}



\subsection{Evaluation Metrics}
Following previous work \cite{gu2021beyond, yang2022towards}, The Macro F1-score (F1) and exact match (EM) are used to evaluate the model's performance. If the predicted answer matches the true answer for each question-answer pair, EM =1. Otherwise, EM = 0. The Macro F1 score is defined as the mean of token-level F1 scores:
\begin{equation}
    Macro\; F1-score = \frac{1}{N} \sum \limits_{i=0}^N F1-score_i
\end{equation}
where $i$ is the token index and \emph{N} is the length of the golden answer.

\section{Results and Discussion}
Our method is applied to both domain-independent QA tasks (\cref{sec:in-domain}) and cross-domain QA tasks (\cref{sec:out-domain}). 

\begin{table*}[h]
\centering
\small
\begin{tabular}{l|llllll}
\toprule
 & \multicolumn{1}{l}{\textbf{S -- \textgreater{}N}} & \multicolumn{1}{l}{\textbf{S -- \textgreater{}C}} & \multicolumn{1}{l}{\textbf{N -- \textgreater{}S}} & \multicolumn{1}{l}{\textbf{N -- \textgreater{}C}} & \multicolumn{1}{l}{\textbf{C -- \textgreater{}S}} & \multicolumn{1}{l}{\textbf{C --\textgreater{} N}} \\ \midrule
\textbf{RoBERTa}        & 37.60                                                    & 66.58                                                      & 49.87                                                    & 44.22                                                       & 19.44                                                      & 7.45                                                        \\
\textbf{RoBERTa+DA} \cite{yue2022synthetic}     & 38.26                                                    & 66.14                                                      & 50.31                                                    & 43.05                                                       & 22.74                                                      & 7.15                                                        \\
\textbf{RoBERTa+P}      & 38.17*                                                   & 66.84                                                      & 50.97*                                                   & 48.37*                                                      & 21.41*                                                     & \textbf{8.64*}                                              \\
\textbf{RoBERTa+LPFT}   & 37.95*                                                   & 66.60                                                      & 50.28*                                                   & 45.86*                                                      & 20.92*                                                     & 7.5                                                         \\
\textbf{RoBERTa+P+LPFT} & \textbf{38.76*}                                          & \textbf{66.86*}                                            & \textbf{52.41*}                                          & \textbf{51.64*}                                             & \textbf{23.02*}                                            & 7.73*                                                       \\ \midrule
\textbf{BART}           & 33.71                                                    & 46.97                                                      & 43.49                                                    & 31.78                                                       & 26.14                                                      & 8.69                                                        \\
\textbf{BART+DA} \cite{yue2022synthetic}        & 35.09                                                    & 55.65                                                      & 44.05                                                    & 33.47                                                       & 26.98                                                      & 9.02                                                        \\
\textbf{BART+P}         & \textbf{36.81*}                                          & \textbf{56.22*}                                            & 43.61*                                                   & 31.91                                                       & 25.96                                                      & 9.26*                                                       \\
\textbf{BART+LPFT}      & 33.29                                                    & 53.29*                                                     & 44.05*                                                   & 31.95                                                       & 26.87*                                                     & 9.49*                                                       \\
\textbf{BART+P+LPFT}    & 35.23*                                                   & 55.86*                                                     & \textbf{44.36*}                                          & \textbf{33.79*}                                             & \textbf{27.61*}                                            & \textbf{9.54*}                                              \\
\bottomrule
\end{tabular}
\caption{OOD test results on SQuAD (S), CausalQA (C), and NewsQA (N) (Setup 2). [S-->N] represents that the model is trained on SQuAD while tested on NewsQA. ``+P'' represents the prompting methods. ``+DA'' represents the Domain Adaptation method \cite{yue2022synthetic}. The proposed method shows statistically significant improvements compared to the baseline model indicated by $*$ using Student T-test (p\textless 0.01, 10-time run).}
\label{tab:squad_newsqa_causalqa}
\end{table*}

\subsection{In-domain Performance}\label{sec:in-domain}
For domain-independent QA, the in-domain performance represents the model performance using the traditional hold-out test, where both the training set and test set come from the whole dataset without splitting domains. The domain-independent results are shown in \emph{Table \ref{tab:1}}, where the Prompt LP-FT method brings performance gain over both the BART model (in average +3.64\% in F1, +4.95\% in EM) and the RoBERTa model (in average +0.42\% in F1, +0.14\% in EM). Taking the BART model as an example, \emph{BART+LP-FT} achieves slightly better performance (+0.57\%) compared with \emph{BART}, which shows the LP-FT method brings limited benefits to the model on the domain-independent QA task. However, \emph{BART+P} (+2.07\%) over \emph{BART} outperforms \emph{BART+LP-FT} (+0.57\%) over \emph{BART}, which shows that the prompt-based method can benefit the model without splitting domains.

In \emph{Table \ref{tab:2}}, the numbers on the diagonal represent the ID performance on each domain, and the values in parentheses below represent the in-domain performance increase brought by our method (in average +2.21\% in F1 and +2.82\% in EM) (\emph{left}). 
Though the performance gain on each domain varies 
, our method consistently improves the performance of in-domain evaluations.

\subsection{Out-of-domain performance}\label{sec:out-domain}
\textbf{Results on CausalQA.} The experiment results of cross-domain CausalQA are shown in $ 3 \times 3$ tables  \emph{Table \ref{tab:2}} where each row represents contrast experiments with the same testing data, and each column represents the model performance on different test sets. The numbers not on the diagonal represent the performance tested on a domain different from the training domain, called OOD test results. Overall, the proposed method benefits the OOD performance by an average of +3.11\% in F1 and 4.76\% in EM on BART and by an average of +2.39\% in F1 and 4.51\% in EM on RoBERTa. For example, by comparison in the same scenario, we find that the improvement on \emph{Consumer-Train/Industrial-Test} on BART (+3.95\%) is more significant than the improvement on \emph{Consumer-Train/Industrial-Test} based on RoBERTa (+2.74\%). Our method brings larger performance gains for generative models (BART) than discriminative models (RoBERTa). These results show that the performance benefits based on discriminative models are less than generative models by using Prompt LP-FT. Intuitively, this can be because the added prompt can be used directly to generate answers as we fine-tune BART in a Seq2Seq manner.

In \emph{Table \ref{tab:2}}, we compare the performance of BART on \emph{Consumer-Train/Consumer-Test} to \emph{Consumer-Train/Industrial-Test}. Our method improves the performance by +0.36\% on the consumer test set and +3.95\% when testing on the samples from the industrial domain, indicating that the proposed method is better for cross-domain generalization. Moreover, the benefit on \emph{Consumer-Train/Technology-Test} (+2.78\%) is relatively small compared to the improvement on \emph{Consumer-Train/Industrial-Test} (+3.95\%). It hints that the same prompt has variant effects on different domains. This can be because different domains have intrinsically different feature distributions. 

\noindent\textbf{OOD Tests Between Different Datasets.} We show the OOD generalization results between different popular datasets in Table \ref{tab:squad_newsqa_causalqa}. It can be seen that the Prompt LP-FT method improves the OOD test performance of RoBERTa on average by \textbf{2.54\%} on three data sets and \textbf{2.60\%} for BART. It is worth noting that our method brings a performance improvement of up to \textbf{7.42\%} (NewsQA--CausalQA) on RoBERTa, while the maximum performance improvement reaches \textbf{8.89\%} (SQuAD--CausalQA) on BART. The result is consistent with the finding in Setup 1 that Prompt LP-FT can benefit generative models more than discriminative ones. 

It is noteworthy that even though our method assumes that no target domain annotations are available (\textbf{zero-shot}), the baseline method using Domain Adaptation (DA) assumes that a small number of target annotations are available (\textbf{few-shot}), our method can consistently achieve better performance than the DA method in all six settings. These results based on the OOD generalization among three datasets suggest that Prompt LP-FT is a highly robust, easy-to-transfer, and convincing method to improve the cross-domain generalization ability of QA models.

\subsection{Discussion}
We provide discussion to understand better the relative contributions of Prompt LP-FT toward performance improvement, including the universality of our method, the ablation study, and case study.

\noindent\textbf{Universality.} The results in \emph{Sec 5.1, 5.2} show that our proposed method improves the OOD generalization performance of various models to varying degrees, with ID performance increasing as well. Experimental results on multiple models demonstrate that our method holds good portability and can benefit variant models, including generative (BART) and discriminative (RoBERTa) models. Results on more backbone models (e.g., T5 and SpanBERT) can be found in \cref{sec:more-result}.

\begin{figure}[t]
    \centering
    \includegraphics[scale=0.36]{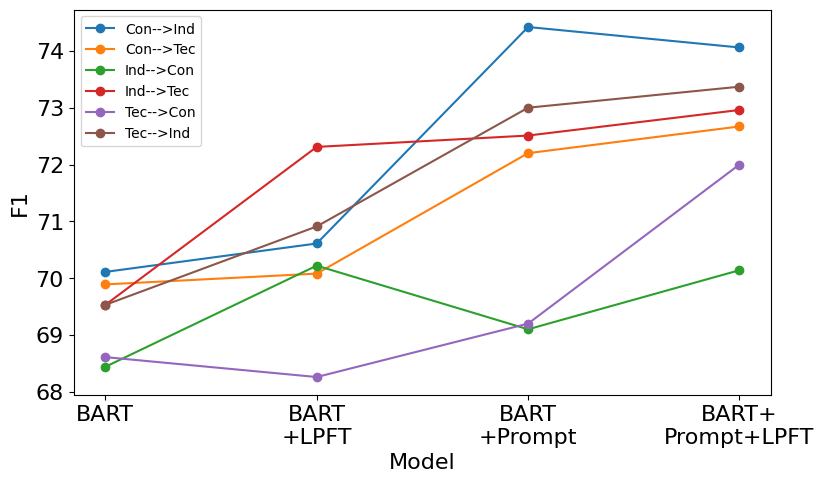}
    \caption{Ablation results based on the BART-base model.}
    \label{fig:ablation_study_BART}
\end{figure}

\noindent\textbf{Ablation Study.} Figure \ref{fig:ablation_study_BART} shows an ablation study of Prompt-based LP-FT. We find that the combination of prompting methods with LP-FT achieves the best performance in four of six settings, illustrating the advantage of prompt-based LP-FT. In addition, BART+Prompt shows an absolute advantage compared to BART+LP-FT, which can be because prompts benefit the cross-domain QA task by introducing more background knowledge than the adjustment of the tuning strategy. The detailed ablation results are shown in Appendix A.5.


\begin{table*}[t]
\resizebox{\linewidth}{!}{
\begin{tabular}{lllll}
\hline
\textbf{Context \& Prompt}                                                                                                                                                                                                                                                                                     & \textbf{Question}                                                                                      & \textbf{Gold Answer}                                                             & \textbf{Baseline Output}                                                                             & \textbf{Our Output}                                                                                                         \\ \hline
\textbf{Predictive Model: SpanBERT-base} & & & & \\ \hline
\resizebox{0.7\linewidth}{!}{\begin{tabular}[c]{@{}l@{}} As Terex has expanded its MP product line, it has captured \\ a larger global market share of the industry, allowing it to\\ \textcolor{teal}{gain greater insight into customer} demand. This may provide \\ revenue synergies in the future...... \\ \textcolor{orange}{Prompt: To answer a "Why useful" question, you need to look} \\ \textcolor{orange}{for "allowing"}\end{tabular}} & \begin{tabular}[c]{@{}l@{}}Why Terex has \\ expanded its MP\\  product line is \\ useful?\end{tabular} & \begin{tabular}[c]{@{}l@{}}\textcolor{teal}{gain greater} \\ \textcolor{teal}{insight into} \\ \textcolor{teal}{customer}\end{tabular} & \begin{tabular}[c]{@{}l@{}} \textcolor{red}{it has captured} \\ \textcolor{red}{a larger global} \\ \textcolor{red}{market share of}\end{tabular} & \begin{tabular}[c]{@{}l@{}}it has captured a larger \\ global market share of \\ the industry, allowing it \\ to \textcolor{teal}{gain greater insight}\\  \textcolor{teal}{into customer}\end{tabular} \\ \hline
\resizebox{0.7\linewidth}{!}{\begin{tabular}[c]{@{}l@{}}However, Avnet's management conceded.....  Suppliers get\\ access to a broad range of customers by using Avnet without \\ having to make significant \textcolor{teal}{investment in sales and engineering} \\ teams. In exchange for these services, Avnet can generate ......\\ \textcolor{orange}{Prompt: The sentiment of this question is positive, you need}\\ \textcolor{orange}{to look for "significant"}\end{tabular}} & \begin{tabular}[c]{@{}l@{}}What will be \\ prevented if \\ suppliers get \\ access to a \\ broad range \\ of customers?\end{tabular} & \begin{tabular}[c]{@{}l@{}}\textcolor{teal}{investment} \\ \textcolor{teal}{in sales and} \\ \textcolor{teal}{engineering}\end{tabular} & \textcolor{red}{using Avnet} & \begin{tabular}[c]{@{}l@{}}using Avnet without \\ having to make \\ significant \textcolor{teal}{investment} \\ \textcolor{teal}{in sales and engineering}\end{tabular}                                 \\ \hline

\textbf{Generative Model: BART-base} & & & & \\ \hline
\resizebox{0.7\linewidth}{!}{\begin{tabular}[c]{@{}l@{}}At the end of 2020, \textcolor{teal}{the store base had grown about 29\% over} \\ \textcolor{teal}{the prior five-year period}, to about 1,920 locations (around \\ 2,100 including Petsense), driving sales and EPS compound \\ annual growth rates over the past three years of 14\% and 27\%, \\ respectively. We forecast that the firm will grow to around .......\\ \textcolor{orange}{Prompt: The entity "EPS" is mentioned in the question. This} \\ \textcolor{orange}{timing "annual" is mentioned in the question.}\end{tabular}} & \begin{tabular}[c]{@{}l@{}} Why sales and \\ EPS compound \\ annual growth \\rates increase?\end{tabular}                                                            & \begin{tabular}[c]{@{}l@{}}\textcolor{teal}{the store base} \\ \textcolor{teal}{had grown} \\ \textcolor{teal}{about 29\%} \\\textcolor{teal}{over the prior}\\ \textcolor{teal}{five-year} \\ \textcolor{teal}{period}\end{tabular}                      & \textcolor{red}{14\% and 27\% }      & \begin{tabular}[c]{@{}l@{}}\textcolor{teal}{the store base had grown} \\ \textcolor{teal}{about 29\% over the} \\ \textcolor{teal}{prior five} years\end{tabular}                                                         \\ \hline
\resizebox{0.7\linewidth}{!}{\begin{tabular}[c]{@{}l@{}}Finally, we view \textcolor{teal}{the likelihood of sustained economic value} \\ \textcolor{teal}{creation as quite high for the restaurant brand}, which finds \\ itself on the leading edge of most of the changes we expect\\ to impact the restaurant industry over the medium to long \\ term. Though Chipotle saw economic value destruction in 201 ......\\ \textcolor{orange}{Prompt: "restaurant industry" is an important phrase. And also} \\ \textcolor{orange}{pay attention to these words:``edge", ``changes"}\end{tabular}}      & \begin{tabular}[c]{@{}l@{}}What will happen \\ if on the leading \\ edge of most of \\ the changes we \\ expect to impact \\ the restaurant \\ industry?\end{tabular} & \begin{tabular}[c]{@{}l@{}}\textcolor{teal}{the likelihood of} \\ \textcolor{teal}{sustained economic}\\  \textcolor{teal}{value creation as} \\ \textcolor{teal}{quite high for the}\\  \textcolor{teal}{restaurant brand}\end{tabular} & \begin{tabular}[c]{@{}l@{}} \textcolor{red}{over the medium} \\ \textcolor{red}{to long term}\end{tabular}                              & \begin{tabular}[c]{@{}l@{}}we view \textcolor{teal}{the likelihood of} \\ \textcolor{teal}{sustained economic value} \\ \textcolor{teal}{creation as quite high for} \\ \textcolor{teal}{the restaurant brand}\end{tabular}                 \\ \hline

\end{tabular}}
\caption{Case study of “Why” and “What-if” questions answering tasks based on the SpanBERT-base and BART-base models. The \textcolor{teal}{Gold Answer} is highlighted using the green text, while the \textcolor{red}{Incorrect Answer} predicted by the baseline method is highlighted by the red text.}
\label{tab:5}
\end{table*}

\noindent\textbf{Case Study} Table \ref{tab:5} presents a case study of four test samples. For each instance, we show the input context, the prompt sentence, and the output predicted by the baseline method and our method (Prompt LP-FT). It can be seen that the gold answers are mostly included in the output of Prompt LP-FT, while the output of baseline models is prone to errors. Specifically, baseline models, including SpanBERT-base and BART-base, tend to output the answers closer to the question in the context instead of observing the whole sentence. For example, for the question ``\textit{What will be prevented if suppliers ... customers?}'', the SpanBERT-base model will output the wrong answer -- ``\textit{using Avnet}'' that is close to the question in the context  -- while the correct answer -- ``\textit{investment in sales and engineering}'' is ignored. These comparisons provide evidence that our method is beneficial in addressing the spurious features of sentence order for QA models. This can be because the well-designed prompt combined with LP-FT helps QA models understand the context better.

\section{Conclusion}

We introduce a zero-shot cross-domain QA task and present a novel Prompt-based LP-FT method by combining prompt with a linear-probing fine-tuning strategy, providing theoretical proof that the proposed method can enhance the model's in-domain performance and out-of-domain generalizability, and empirically showing that the Prompt LP-FT method consistently benefits the QA models. Experimental results show that (1) current methods still have a lag much behind human-level towards the cross-domain QA generalization; (2) our method brings larger performance gains for generative models than discriminative models; (3) the use of the prompt-based LP-FT in other NLP tasks is worth trying. Meanwhile, the emergent ability of LLMs will definitely decrease the challenge of the current cross-domain QA setting. Designing challenging datasets of cross-domain QA towards LLMs should be paid more attention in the future.

\section*{Limitation}
Our method has a few limitations which may inspire future work. First, the prompt templates are manually designed, although we've introduced the rules and intuitions used in our implementation. Second, the proposed method may have low scalability to long text. Because we add the prompt at the end of the context, the prompt would be truncated if the context itself exceeds the maximum acceptable token length of the model.

\section*{Ethics Statement}
This paper honors the ACL Code of Ethics. Public available datasets are used to establish our results. No private data and crowd-sourcing work are used to produce predictions. The code and data are open-sourced under the CC-BY-NC-SA license.

\section*{Acknowledgement}
This publication has emanated from research conducted with the financial support of Science Foundation Ireland under Grant number 18/CRT/6183, the financial support of the Pioneer and ``Leading Goose" R\&D Program of Zhejiang under Grant Number 2022SDXHDX0003 and the 72nd round of the Chinese Post-doctoral Science Foundation project 2022M722836. For the purpose of Open Access, the author has applied a CC BY public copyright licence to any Author Accepted Manuscript version arising from this submission. Yue Zhang is the corresponding author.



\appendix
\section{Appendix}

\subsection{Template Comparison}\label{sec:template-comparison}

As we see in Table \ref{tab:proper_sen}, changing "But" in the template to "And" alters the logical relationship between the preceding and following sentences, which had an impact of more than 1\% on the performance.


\subsection{Template Engineering}\label{sec:template-engineering}

The main objective of applying prompt templates is to enhance the model's out-of-domain performance by extracting invariant features between different domain questions. Therefore, the first rule is that a designed template should avoid containing domain-related information. For example, "This [health] company [Hologic] is mentioned in the question." should not be an ideal template because it involves extra domain information that Hologic is a health company. 




\begin{figure}[t]
    \centering
    \includegraphics[scale=0.4]{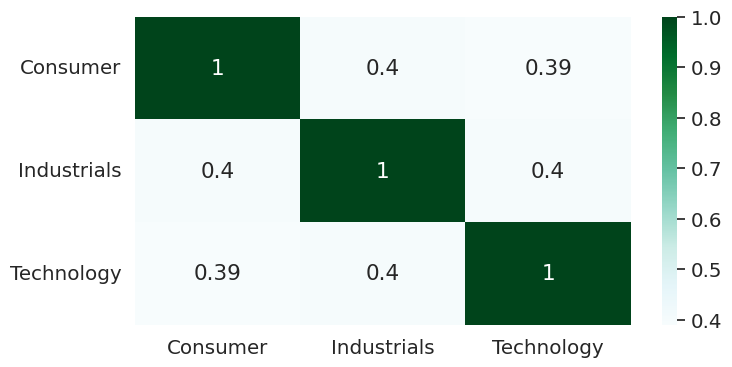}
    \caption{The word overlap between different datasets considered by the CasualQA task.}
    \label{fig:causalqa_Overlap}
\end{figure}

\begin{figure}[t]
    \centering
    \includegraphics[scale=0.4]{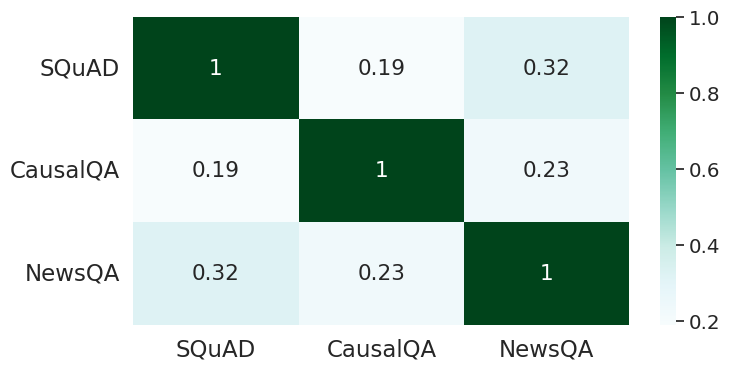}
    \caption{The word overlap between SQuAD, CausalQA and NewsQA datasets.}
    \label{fig:squad_newsqa_Overlap}
\end{figure}

\begin{table}[t]
\begin{tabular}{|l|l|l|ll}
\cline{1-3}
            & Template                                                                                                                             & F1 &  &  \\ \cline{1-3}
Baseline    & None                                                                                                                                 & 70.29   &  &  \\ \cline{1-3}
Experiment1 & \begin{tabular}[c]{@{}l@{}}"There is no important\\ phrase in this query. \\ But also pay attention\\ to these words: \_\_"\end{tabular} & 69.57   &  &  \\ \cline{1-3}
Experiment2 & \begin{tabular}[c]{@{}l@{}}"There is no important\\ phrase in this query. \\ And also pay attention\\ to these words: \_\_"\end{tabular} & 70.84   &  &  \\ \cline{1-3}
\end{tabular}
\caption{The effect of using an improper word in a template.}
\label{tab:proper_sen}
\end{table}

Second, a template should be a proper English sentence with correct spelling, no grammar mistakes, and proper semantic meaning. Our experiment shows that one wrong word in a template may cause significant performance variation (see Appendix A.1).

Third, since templates are concatenated at the end of the context, templates cannot be too long. If a template has almost the same length as the context or even longer, it will double the amount of input data and thus increase the computational cost of the model; more importantly, it may deprive the leader role of the context, which may make the model too generalized to capture the answers. 

Fourth, there are two main varieties of prompt templates: \textit{cloze prompt} and \textit{prefix prompt}.\cite{liu2021promptsurvey}  Cloze prompt fill in the blanks of a textual string, and the prefix prompt continue a string prefix. Instead of using only one type, we include both variants of templates in the designed four prompt templates. \\


According to these rules, we design four types of templates, of which each type has different sentence patterns. Template generalization is modulized as a two-step process: \textbf{1)} generating the prompt words, and \textbf{2)} filling in the blanks \cite{liu2021promptsurvey}.

\subsection{Template Designing}\label{sec:template-designing}
\textbf{Sentiment Templates} 
Assume that a person unfamiliar with the restaurant industry tries to answer the question, \emph{``Why the global restaurant sector has come under pressure?''}. This person can easily find that this question concerns the factors that adversely affect restaurants even without industry knowledge. Therefore, looking for negative words from the context, like \emph{destroyed}, \emph{restricted} etc., may help to locate the correct answer. Based on the intuition above, we implement a sentiment analysis framework \footnote{implemented using the NLTK module} to give each question and each word in the answer sentence a sentiment score. Afterwards, the highest positive or negative scores are selected to be used as the prompt words. Second, the sentient of the question and the prompt words are filled in the blanks of sentiment templates.

\textbf{Named Entity Templates} Unique entities mentioned in a question could hint at answering the question. Hence, a named entity recognition framework is applied to each question. We intend to recognize five types of entities mentioned in the question: Person, Organization, Location, Country, and Date. Entities not included in the five types are assigned as ``Other'' entities. Step two fills the recognized entities in the blanks as prompt words. 


\textbf{Phrase Template} Phrases are usually the question subject, potentially valuable in locating the correct answer. A simple strategy is designed to find out the phrases composed of an adjective(s) and noun(s). For example, ``hybrid environments'', ``software-as-a-service applications'', and ``remote access'' are phrases in a question. These phrases are selected as prompt words and filled in the blanks in step two. 


\subsection{Word Overlap between Datasets}\label{sec:word-overlap}
Fig \ref{fig:causalqa_Overlap} and \ref{fig:squad_newsqa_Overlap} show the word overlap percentage between different domains of the CausalQA dataset, and also on datasets from different domains, i.e., between the SQuAD, CausalQA and NewsQA datasets.

\subsection{Experiment Results on Other Models}\label{sec:more-result}

\begin{table}[t]
\centering
\small
\begin{tabular}{lllll}
\hline
\textbf{}               & \multicolumn{2}{c}{\textbf{Dev}} & \multicolumn{2}{c}{\textbf{Test}} \\
\textbf{Methods}        & F1              & EM             & F1              & EM              \\ \hline
\multicolumn{1}{l|}{\textbf{SpanBERT-base}}  & 84.77           & 62.62          & 84.85           & 64.04           \\ 
\multicolumn{1}{l|}{\textbf{SpanBERT-large}} & 85.40    &   61.41          &    85.53    &  62.26        \\ \hline
\end{tabular}
\caption{Domain-independent QA results of SpanBERT-base and SpanBERT-large model.}
\label{spanbert}
\end{table}

\begin{table}[t]
\resizebox{\linewidth}{!}{
\begin{tabular}{lllllll}
\hline
\textbf{Domain} & \multicolumn{2}{l}{\textbf{Con}} & \multicolumn{2}{l}{\textbf{Ind}} & \multicolumn{2}{l}{\textbf{Tech}} \\
                & F1                     & EM           & F1                      & EM            & F1                      & EM            \\ \hline
\multicolumn{1}{l|}{\textbf{Con}}    & \textbf{85.84}         & 60.64        & 84.80                    & 56.01         & 85.54                   & 55.09         \\ 
\multicolumn{1}{l|}{\textbf{Ind}}  & 85.76                  & 66.66        & \textbf{85.84}          & 67.21         & 85.34                   & 65.57         \\ 
\multicolumn{1}{l|}{\textbf{Tech}}   & 80.24                  & 58.08        & 81.51                   & 58.58         & \textbf{81.98}          & 58.08         \\ \hline
\end{tabular}}
\caption{Out-of-domain test results of SpanBERT-base.}
\label{tab:span_ood}
\end{table}

On both domain-independent QA and cross-domain QA tasks, the SpanBERT model achieves state-of-the-art performance. \emph{Table \ref{spanbert}} shows the domain-independent QA result of SpanBERT-base and SpanBERT-large model, which also provides evidence that the proposed method works on the large model which can achieve better results than it on the base model. Tab\ref{tab:span_ood} shows the result of the Span-BERT OOD test.

Tab \ref{tab:t5} shows the cross-domain QA experiment results on T5-base. We show that our method can significantly improve the cross-domain QA performance compared to the standard fine-tuning results based on the CausalQA dataset.

\begin{figure}[t]
    \centering
    \includegraphics[scale=0.37]{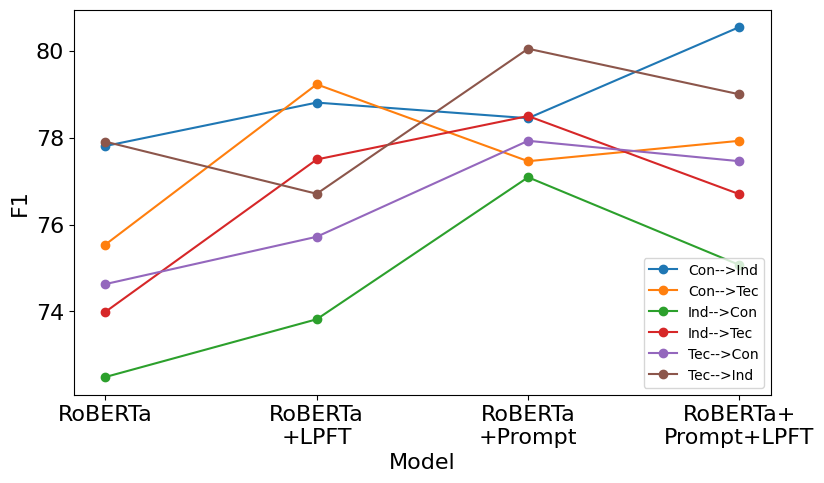}
    \caption{Ablation study results based on the RoBERTa-base model for cross-domain QA.}
    \label{fig:ablation_study_RoBERTa}
\end{figure}

Tab \ref{tab:RoBERTa_ablation}, \ref{tab:BART_ablation} and Fig \ref{fig:ablation_study_RoBERTa} are the ablation study results on RoBERTa and BART models for cross-domain QA task on the CausalQA dataset.

\begin{table}[t]
\resizebox{\linewidth}{!}{
\begin{tabular}{llllllll}
\hline
\multicolumn{2}{l}{\textbf{Train/Test}}                        & \multicolumn{2}{l}{\textbf{Consumer}} & \multicolumn{2}{l}{\textbf{Industrial}} & \multicolumn{2}{l}{\textbf{Technology}} \\
                                                     &                           & F1                    & EM            & F1                     & EM             & F1                     & EM             \\ \hline
\multicolumn{1}{l|}{\multirow{2}{*}{\textbf{Con}}}    & \multicolumn{1}{l|}{Ori}  & \textbf{59.30}        & 15.74         & 56.69                  & 18.60          & 56.13                  & 17.67          \\ \cline{2-2}
\multicolumn{1}{l|}{}                                & \multicolumn{1}{l|}{Ours} & (+2.19)               & (+3.87)       & (+1.79)                & (+0.84)        & (+2.24)                & (+1.77)        \\ \cline{1-2}
\multicolumn{1}{l|}{\multirow{2}{*}{\textbf{Ind}}} & \multicolumn{1}{l|}{Ori}  & 59.85                 & 24.72         & \textbf{61.41}         & 24.03          & 59.94                  & 25.27          \\ \cline{2-2}
\multicolumn{1}{l|}{}                                & \multicolumn{1}{l|}{Ours} & (+2.52)               & (-0.68)       & (+0.06)                & (-3.81)        & (+1.64)                & (-4.51)        \\ \cline{1-2}
\multicolumn{1}{l|}{\multirow{2}{*}{\textbf{Tech}}}  & \multicolumn{1}{l|}{Ori}  & 55.30                 & 17.85         & 54.41                  & 16.83          & \textbf{58.76}         & 20.20          \\ \cline{2-2}
\multicolumn{1}{l|}{}                                & \multicolumn{1}{l|}{Ours} & (+3.70)               & (+5.38)       & (+4.83)                & (+5.39)        & (+0.53)                & (+2.02)        \\ \hline
\end{tabular}}
\caption{Out-of-domain test results of the T5-base model. Numbers in brackets represent the performance improved by our method.}
\label{tab:t5}
\end{table}

\begin{table*}[t]
\centering
\small
\begin{tabular}{llllllll}
\hline
                                                                                                                      &              & \multicolumn{2}{l}{\textbf{Consumer}} & \multicolumn{2}{l}{\textbf{Industrial}} & \multicolumn{2}{l}{\textbf{Technology}} \\
\textbf{Methods}                                                                                                      &              & F1                & EM                & F1                 & EM                 & F1                 & EM                 \\ \hline
\multirow{3}{*}{\textbf{\begin{tabular}[c]{@{}l@{}}Baseline:\\ RoBERTa\end{tabular}}}            & \textbf{Con} & 78.20              & 51.38             & 72.49              & 47.68              & 74.63              & 49.07              \\
                                                                                                                      & \textbf{Ind} & 77.81             & 49.45             & 80.05              & 58.46              & 77.91              & 48.35              \\
                                                                                                                      & \textbf{Tec} & 75.54             & 55.05             & 73.99              & 48.98              & 76.49              & 54.04              \\ \hline
\multirow{3}{*}{\textbf{RoBERTa + LP-FT}}                                                        & \textbf{Con} & 75.09             & 47.68             & 73.82              & 46.75              & 75.72              & 50.00                 \\
                                                                                                                      & \textbf{Ind} & 78.81             & 48.90              & \textbf{81.01*}    & 51.10               & 76.71              & 50.00                 \\
                                                                                                                      & \textbf{Tec} & 79.23             & \textbf{56.06*}   & 77.50               & 55.05              & 78.22              & 53.53              \\ \hline
\multirow{3}{*}{\textbf{\begin{tabular}[c]{@{}l@{}}RoBERTa + Prompt\end{tabular}}}       & \textbf{Con} & \textbf{78.97*}   & 53.24             & \textbf{77.09*}    & \textbf{50.00*}       & \textbf{77.93*}    & \textbf{51.85*}    \\
                                                                                                                      & \textbf{Ind} & 78.45             & 56.28             & 80.05              & \textbf{61.74*}    & \textbf{80.05*}    & \textbf{57.92*}    \\
                                                                                                                      & \textbf{Tec} & 77.46             & 54.54             & \textbf{78.50*}     & \textbf{57.58*}    & \textbf{80.62*}    & \textbf{58.08*}    \\ \hline
\multirow{3}{*}{\textbf{\begin{tabular}[c]{@{}l@{}}RoBERTa + LP-FT + Prompt\end{tabular}}} & \textbf{Con} & 78.39             & \textbf{53.24*}   & 75.07              & 48.61              & 77.46              & 49.54              \\
                                                                                                                      & \textbf{Ind} & \textbf{80.55*}   & \textbf{59.02*}   & 80.70               & 60.66              & 79.00                 & 57.37              \\
                                                                                                                      & \textbf{Tec} & \textbf{77.93*}   & 55.05             & 76.70               & 56.06              & 76.76              & 53.54              \\ \hline
\end{tabular}
\caption{Ablation study results based on the RoBERTa-base model. Also, the results are averaged by 10 repeated experiments. The statistically significant performance improvements of our proposed method compared to the baseline model are indicated by \(*\) based on the T-test ($P < 0.01$).}
\label{tab:RoBERTa_ablation}
\end{table*}

\begin{table*}[h]
\centering
\small
\begin{tabular}{llllllll}
\hline
                                                                                                                      &              & \multicolumn{2}{l}{\textbf{Consumer}} & \multicolumn{2}{l}{\textbf{Industrial}} & \multicolumn{2}{l}{\textbf{Technology}} \\
\textbf{Methods}                                                                                                      &              & F1                & EM                & F1                 & EM                 & F1                 & EM                 \\ \hline
\multirow{3}{*}{\textbf{\begin{tabular}[c]{@{}l@{}}Baseline:\\ BART
\end{tabular}}}            & \textbf{Con} & 70.29             & 24.53             & 68.44              & 24.07              & 68.61              & 23.61              \\
                                                                                                                      & \textbf{Ind} & 70.11             & 31.31             & 72.53              & 32.41              & 69.53              & 27.77              \\
                                                                                                                      & \textbf{Tec} & 69.89             & 30.30              & 69.53              & 27.77              & 71.79              & 33.83              \\ \hline
\multirow{3}{*}{\textbf{BART + LP-FT}}                                                        & \textbf{Con} & \textbf{70.81*}   & 24.07             & \textbf{70.22*}    & 24.53              & 68.26              & 23.61              \\
                                                                                                                      & \textbf{Ind} & 70.61             & 30.77             & 73.39              & 31.87              & 70.91              & 28.57              \\
                                                                                                                      & \textbf{Tec} & 71.08             & \textbf{35.35}    & 72.31              & 30.30               & 72.37              & 33.33              \\ \hline
\multirow{3}{*}{\textbf{\begin{tabular}[c]{@{}l@{}}BART + Prompt\end{tabular}}}       & \textbf{Con} & 70.41             & \textbf{27.31*}   & 69.10               & 23.15              & 69.20               & 23.61              \\
                                                                                                                      & \textbf{Ind} & \textbf{74.42*}   & \textbf{42.07*}   & 76.27              & 37.16              & 73.00                 & 34.43              \\
                                                                                                                      & \textbf{Tec} & 72.20              & 32.32             &72.51    & 30.3               & 73.32              & \textbf{35.35*}    \\ \hline
\multirow{3}{*}{\textbf{\begin{tabular}[c]{@{}l@{}}BART + LP-FT + Prompt\end{tabular}}} & \textbf{Con} & 70.65             & 25.01                & 70.14              & \textbf{26.85*}    & \textbf{72.00*}       & \textbf{28.24*}    \\
                                                                                                                      & \textbf{Ind} & 74.06             & 36.06             & \textbf{76.58*}    & \textbf{39.89*}    & \textbf{73.37*}    & \textbf{37.70*}     \\
                                                                                                                      & \textbf{Tec} & \textbf{72.67*}   & 33.33             & \textbf{72.96*}            & \textbf{31.31*}    & \textbf{74.02*}    & 34.34              \\ \hline
\end{tabular}
\caption{Ablation study results based on the BART-base model for cross-domain QA. The results are averaged by 10 repeated experiments. The statistically significant performance improvements of our proposed method compared to the baseline model are indicated by \(*\) based on the T-test ($P < 0.01$).}
\label{tab:BART_ablation}
\end{table*}

\subsection{Details of experimental results}

The experiment is conducted on a GTX 3090 TI with 24GB graphics RAM size. The average training time for each model on the domain-independent QA task is around 2.5 hours, and on the cross-domain QA task is around 30 minutes on CausalQA dataset. On SQuAD and NewsQA dataset, the average training time for each model is around 3 hours. For each experiment setting, we run 10 repeated experiments and report the average results. The model name indicates the base model is no size specification, e.g. "BART+P" indicates the BART-base model plus the prompting method. We also implemented large models to prove the effectiveness of the proposed models. 

For the hyperparameter tuning, we split the whole dataset into train/validation/test sets on the domain-independent QA task and use the validation set for hyperparameter tuning. On the cross-domain QA task, we split the dataset of each domain into train/validation/test sets and use the validation set that comes from the same domain with the training set for hyperparameter tuning. The criterion used to select the hyperparameter is the F1 on the validation set. We first select a series of candidate values of a hyperparameter through uniform sampling from a reasonable range, then select the value that achieves the best F1 on the validation set. Three repeated trials decide the value of a hyperparameter. For example, we give the best-performing RoBERTa-base model configuration on \emph{Consumer-Train/Technology-Test} experiment as follows: the learning rate for linear-probing is $10^{-6}$, the number of epochs for linear probing is 3, the learning rate for fine-tuning is $10^{-5}$, the training batch size is 4, parameters are updated every 8 batches, and the number of epochs for fine-tuning is 14. 

\end{document}